\documentclass[a4paper,twocolumn]{paper}
\usepackage{geometry}
\usepackage{graphicx}
\usepackage[font=small,labelfont=bf,tableposition=top]{caption}

\author{
	\hfill
	\hspace{2cm}
	Justus Thies$^1$\hfill
	Michael Zollh{\"o}fer$^2$\hfill
	Matthias Nie{\ss}ner$^1$\hfill
	\\
	\hspace{2cm}
	$^1$Technical University of Munich\hspace{1.1cm}
	$^2$Stanford University
	\vspace{-0.3cm}
}

\title{IMU2Face: Real-time Gesture-driven Facial Reenactment\vspace{-0.5cm}}

\begin{document}
	\twocolumn[
	\maketitle
	

	\hrule 
	
	\begin{abstract} {
			
			{\centering
				\includegraphics[width=\textwidth]{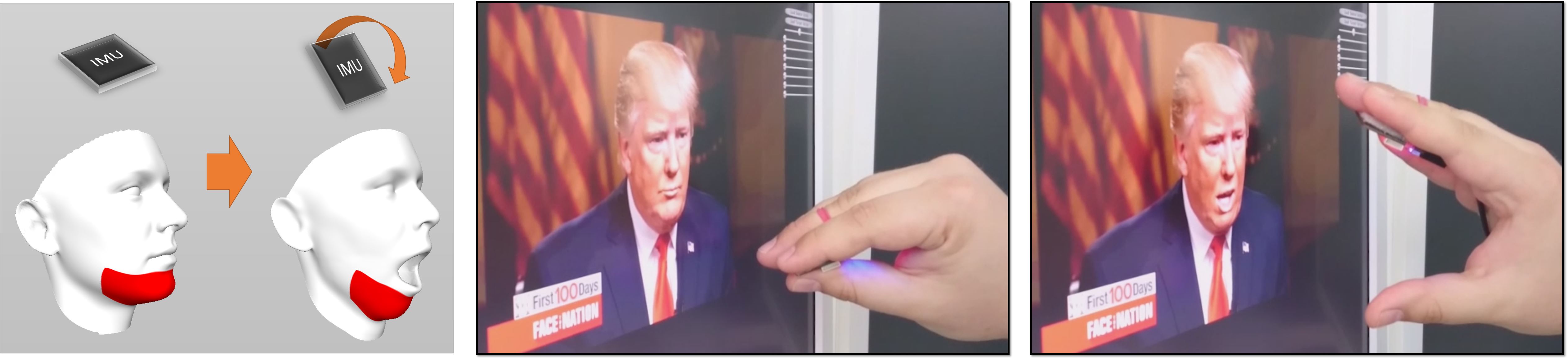}
				\par
			}

			We present IMU2Face, a gesture-driven facial reenactment system.
			To this end, we combine recent advances in facial motion capture and inertial measurement units (IMUs) to control the facial expressions of a person in a target video based on intuitive hand gestures.
			IMUs are omnipresent, since modern smart-phones, smart-watches and drones integrate such sensors; e.g., for changing the orientation of the screen content, counting steps, or for flight stabilization.
			Face tracking and reenactment is based on the state-of-the-art real-time Face2Face~\cite{Face2Face} facial reenactment system.
			Instead of transferring facial expressions from a source to a target actor, we employ an IMU to track the hand gestures of a source actor and use its orientation to modify the target actor's expressions; see the supplemental video \url{https://youtu.be/UXGodiDAqiE}.
			} \end{abstract}
	\vspace{-0.2cm}
	\hrule
	\bigskip
	\vspace{-0.2cm}
	]

   %

\section{Related Work}
\vspace{-0.3cm}
This project has been inspired by the Sparse Inertial Poser~\cite{SIP} body tracking approach.
This approach employs a small set of six body-mounted IMUs to reconstruct the full body motion of an actor.
The first real-time facial reenactment approach has been introduced by Thies et al.~\cite{FaceRGBD}.
In this approach, an RGB-D sensor, e.g., the Microsoft Kinect, is used to track and reconstruct human faces enabling live expression transfer from one person to another.
In a follow-up work, the Face2Face \cite{Face2Face} approach, for the first time, enabled real-time reenactment of standard in-the-wild color videos; i.e., downloaded from Youtube.
Recently, the FaceVR \cite{FaceVR} approach demonstrated self-reenactment for head mounted display (HMD) removal to enable teleconferencing in VR, where participants want to see each other without an HMD occluding half of the face.

\vspace{-0.3cm}
\section{Method Overview}
\vspace{-0.3cm}
Given a video of a target actor, we first analyze the facial geometry and expressions using the Face2Face \cite{Face2Face} approach.
Based on the obtained 3D actor model, the facial expressions in the target video can be edited.
Instead of transferring facial expressions from a source video, we use an IMU that is attached to the source actor's hand to control the target actor; i.e., the jaw motion.
For details on facial tracking, we refer to the Face2Face publication \cite{Face2Face}.
In the following, we describe the reconstruction of the expression parameters based on an IMU.

\vspace{-0.3cm}
\section{Motion Transfer}
\vspace{-0.3cm}
The target actor's expressions are modified to visually match the hand gesture of the source actor.
Given the orientation of the IMU, we compute the expression blendshape weights that best explain such an orientation of the jaw bone.
To this end, we minimize the deformation transfer (DT) energy \cite{DT}.
We measure the DT energy over the jaw region (see figure, left).
The target deformation gradients in the jaw region are defined by the relative orientation of the IMU sensor.
Thus, as an initialization, a reference pose of the IMU sensor has to be defined.
The resulting linear system is solved in real-time for every frame.
We employ an RC drone flight controller that integrates and filters the IMU data at high rates.
We use the open flight controller software BetaFlight\footnote{\url{https://github.com/betaflight/betaflight/releases}} to stream the IMU data to a desktop PC.

\vspace{-0.3cm}
\section{Conclusion}
\vspace{-0.3cm}
With IMU2Face, we demonstrated a novel method to adapt the expressions of a target actor based on intuitive hand gestures.
This show-casts the easy applicability of IMUs for real time tracking.
We believe similar strategies could be employed to drive other actions.

\bibliographystyle{plain}
{\footnotesize
	\bibliography{mainbib}}

\begin{thebibliography}{1}

\bibitem{DT}
R.~W. Sumner and J.~Popovi\'{c}.
\newblock Deformation transfer for triangle meshes.
\newblock In {\em ACM SIGGRAPH 2004 Papers}, SIGGRAPH '04, pages 399--405, New
  York, NY, USA, 2004. ACM.

\bibitem{FaceRGBD}
J.~Thies, M.~Zollh{\"o}fer, M.~Nie{ss}ner, L.~Valgaerts, M.~Stamminger, and
  C.~Theobalt.
\newblock Real-time expression transfer for facial reenactment.
\newblock {\em ACM Transactions on Graphics (TOG)}, 34(6), 2015.

\bibitem{Face2Face}
J.~Thies, M.~Zollh{\"o}fer, M.~Stamminger, C.~Theobalt, and M.~Nie{\ss}ner.
\newblock {Face2Face: Real-time Face Capture and Reenactment of RGB Videos}.
\newblock In {\em Proc. Computer Vision and Pattern Recognition (CVPR), IEEE},
  2016.

\bibitem{FaceVR}
J.~Thies, M.~Zollh{\"o}fer, M.~Stamminger, C.~Theobalt, and M.~Nie{ss}ner.
\newblock Facevr: Real-time facial reenactment and eye gaze control in virtual
  reality.
\newblock 2016.

\bibitem{SIP}
T.~{von Marcard}, B.~Rosenhahn, M.~Black, and G.~Pons-Moll.
\newblock Sparse inertial poser: Automatic 3d human pose estimation from sparse
  imus.
\newblock {\em Computer Graphics Forum 36(2), Proceedings of the 38th Annual
  Conference of the European Association for Computer Graphics (Eurographics)},
  2017.

\end{thebibliography}

\end{document}